%% file: main.tex
\documentclass[10pt,twocolumn,letterpaper]{article}

\usepackage{iccv} 

\definecolor{cvprblue}{rgb}{0.21,0.49,0.74}
\usepackage[pagebackref,breaklinks,colorlinks,allcolors=cvprblue]{hyperref}

\usepackage{booktabs}       
\usepackage{amsfonts}       
\usepackage{nicefrac}       
\usepackage{microtype}      
\usepackage[inline]{enumitem}

\usepackage{sidecap,wasysym,array,tabularx,caption}
\usepackage[ruled]{algorithm}
\usepackage{algpseudocode}
\usepackage{todonotes}
\usepackage{multirow}

\usepackage{pifont}
\usepackage{xcolor}
\usepackage[capitalize]{cleveref}
\usepackage{mathtools}
\usepackage{tabularx}

\usepackage{booktabs} 
\usepackage{color} 
\usepackage{wrapfig}

\usepackage{arydshln}

\newcommand{\greencheck}{{\color{green}\ding{51}}} 
\newcommand{\redx}{{\color{red}\ding{55}}} 

\title{DDIL: Diversity Enhancing Diffusion Distillation with Imitation Learning}

\author{
    \textbf{Risheek ~Garrepalli} \quad \textbf{Shweta Mahajan} \quad \textbf{Munawar Hayat} \quad \textbf{Fatih Porikli} \\
    Qualcomm AI Research\thanks{Qualcomm AI Research is an initiative of Qualcomm Technologies, Inc.}\\
  \texttt{\{ rgarrepa, shwemaha, mhayat, fporikli\}@qti.qualcomm.com} \\
}

\begin{document}
\maketitle

\input{sections/0_abstract}  
\vspace{-14pt}
\input{sections/intro}    
\input{sections/relatedwork}

\input{sections/background}

\input{sections/method_ddil}
\input{sections/experiments}

\section{Conclusion}


This work introduces DDIL, a novel framework for distilling diffusion models that addresses the challenge of covariate shift while preserving the marginal data distribution. Integrating DDIL with established distillation techniques like LCM, DMD2 and PD consistently yields quantitative and qualitative improvement without loss of diversity demonstrating generalization across various methods. This is particularly noteworthy given the already high performance of distilled models, which often rival their teacher models with limited headroom for improvement. 

Furthermore, we also show that integrating DDIL within the DMD2 framework enhances training stability, achieves better diversity than the SSD1B teacher model, even with a significantly smaller batch size (7 compared to 128 in the original DMD2 training setup). This highlights computational efficiency, diversity preserving and practical usefulness of DDIL across distillation methods.




\clearpage 

{
    \small
    \bibliographystyle{ieeenat_fullname}
    \bibliography{main}
}

\newpage

\input{sections/supplementary}



\end{document}

%% file: sections/0_abstract.tex
\begin{abstract}


Distilling Diffusion models has enabled practical adoption of diffusion models by reducing the number of iterations but distillation techniques often could suffer from lack of diversity, quality, etc.
In this work we enhance training distribution for distilling diffusion models by training on both data distribution (forward diffusion) and student induced distributions (reverse process at inference). We formulate diffusion distillation within imitation learning (\textit{DDIL}) framework and identify co-variate shift i.e., difference in intermediate marginal distributions between training and inference leading to poor performance of multi-step distilled models from compounding error at inference time.

Training on data distribution helps to diversify the generations by \textit{preserving marginal data distribution} and training on student distribution addresses compounding error by \textit{correcting covariate shift}. In addition, we adopt reflected diffusion formulation for distillation and demonstrate improved performance, stable training across distillation methods resulting in \textbf{18.73 FID@30k on SSD1B with just batch size `7' and 40K} updates! We show that DDIL consistently improves on baseline algorithms of Distribution Matching Distillation \textit{(DMD2)},  consistency distillation \textit{(LCM)} and progressive distillation \textit{(PD)}.
\end{abstract}

%% file: sections/intro.tex
\section{Introduction}\label{sec:introduction}


\begin{figure}
  \centering

    \includegraphics[width=\linewidth]{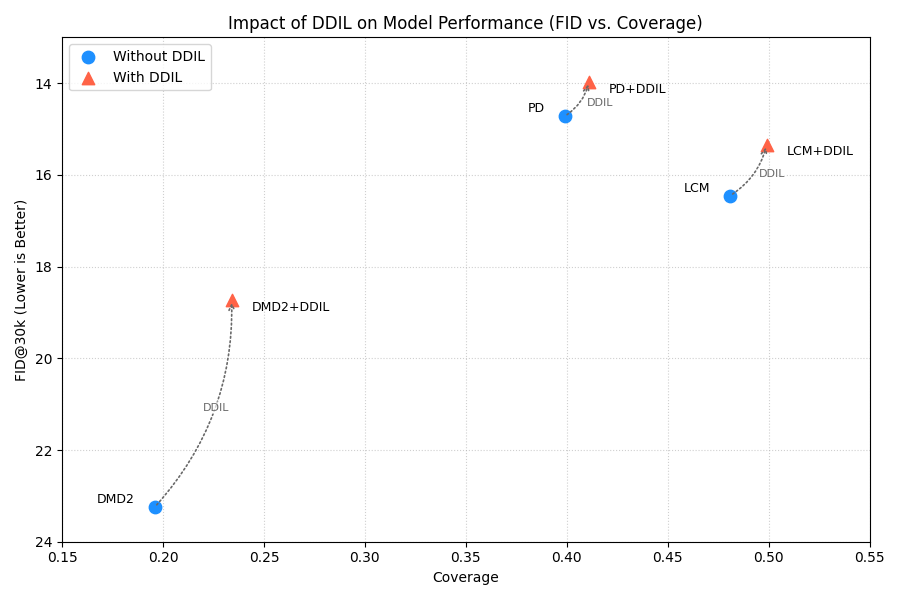}

  \caption{
  DDIL consistently improves both sample quality (FID@30k) and diversity(Coverage) across distillation approaches. Performance shown for DMD2 applied to SSD1B, and Consistency/Progressive Distillation applied to SDv1.5. Coverage measures the extent to which the generated samples span the real data manifold.
  }
  \label{fig:diversity}
  \vspace{-15pt}
\end{figure}

\begin{figure*}
  \centering
    
    

    \includegraphics[width=\linewidth,trim=0cm 2cm 0cm 1cm, clip]{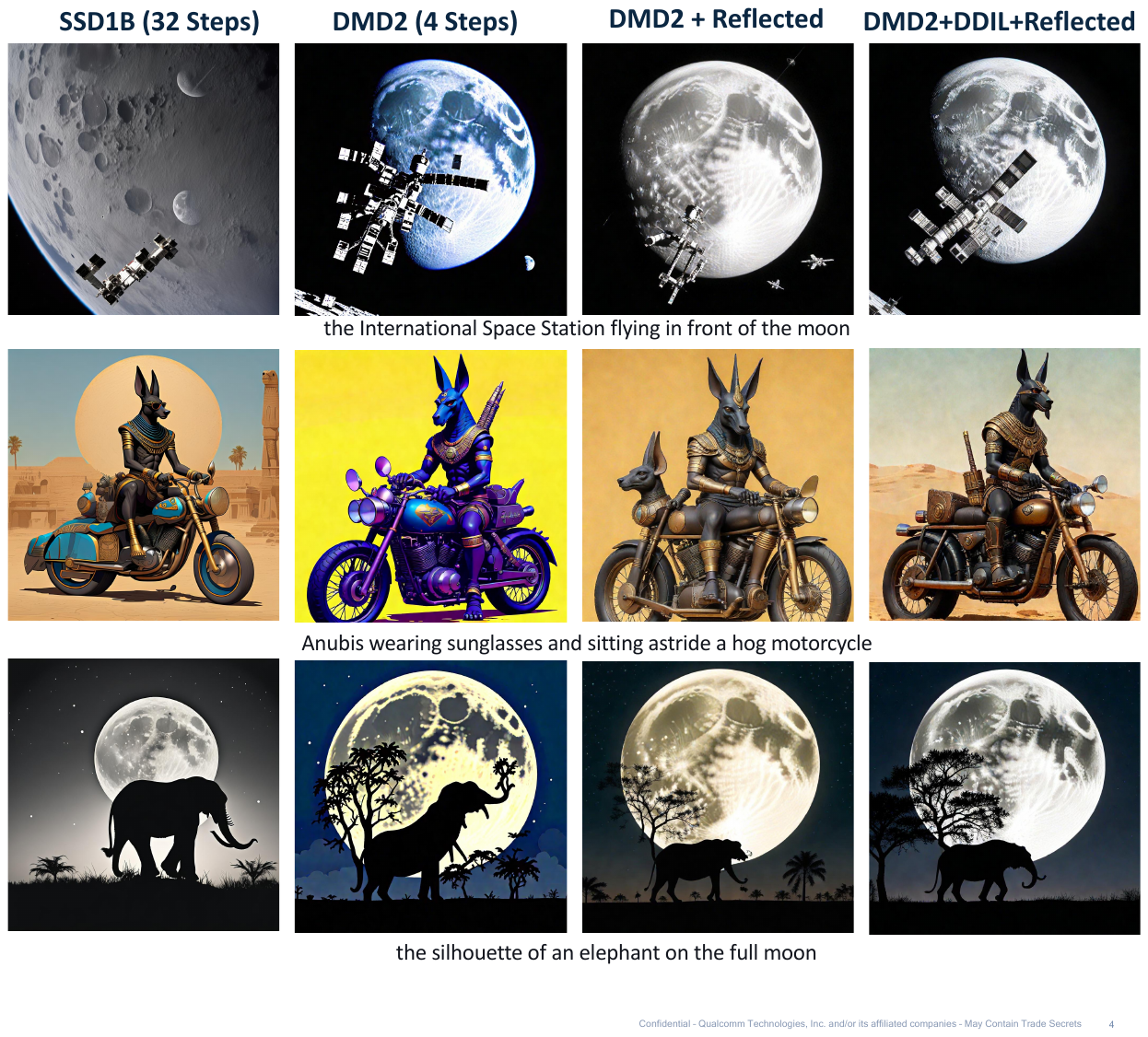}
    
  \caption{Qualitative comparison of images generated with different distillation techniques.We can observe more coherent structure with DDIL compared to baselines DMD2 even with thresholding e.g., space station structure or motorcycle structure. All distilled models are trained on same dataset, batch size and evaluated on same seed and hence generations share characteristics}
  \label{fig:image1}
  \vspace{-15pt}
\end{figure*}

Diffusion models, while capable of producing high-quality images, suffer from slow sampling times due to their iterative denoising process. To address this, distillation techniques have been proposed to reduce number of denoising steps. These techniques can be broadly categorized into trajectory-level \citep{luo2023latent, meng2023distillation, salimans2021progressive,song2023consistency}  and distribution-matching approaches \citep{yin2023one, yin2024improveddistributionmatchingdistillation, luo2024diff, sauer2023adversarial} . While the former focuses on preserving the teacher's trajectory at a per-sample level, the latter matches the marginal distribution.

Multi-step student models offer a promising approach in balancing quality and computational efficiency.
However, multi-step student model also suffer from `covariate shift', i.e., difference in real and student's assumed marginal distributions at intermediate time-steps or noise-levels. This is largely due to training vs inference mismatch in diffusion models and the effect on performance become more pronounced in few-step regime of distilled models.
Recent works \cite{kohler2024imagine, yin2024improveddistributionmatchingdistillation} 
address this issue by distilling on backward trajectories but ignore diversity of generations. There is significant room for improvement on diversity-preserving distillation and need for consistent reporting of relevant metrics and identify better metrics w.r.t diversity.

In this work, we formulate diffusion distillation within the imitation learning (DDIL) framework and address covariate shift and also identify implicit assumptions in training distributions across different distillation techniques potentially leading to loss of diversity of distilled models.We achieve this by incorporating both the data distribution (forward diffusion) and that student's predictive distribution (backward trajectory at inference time). 

This approach combines the benefits of \textit{(1) Preserving Marginal Data Distribution:} Training on the data distribution ensures the student model maintains the inherent statistical properties of the original data, and \textit{(2) Correcting Covariate Shift:} Training on backward trajectories enables the student model to identify and adapt to covariate shifts, thereby improving the accuracy of score estimates, particularly in few-step settings. We illustrate instantiation of DDIL framework in context of progressive distillation in \Cref{fig:rollouts}. To this end, we make the following contributions:

\begin{figure*}
    \centering
    \includegraphics[width=1.0\textwidth]{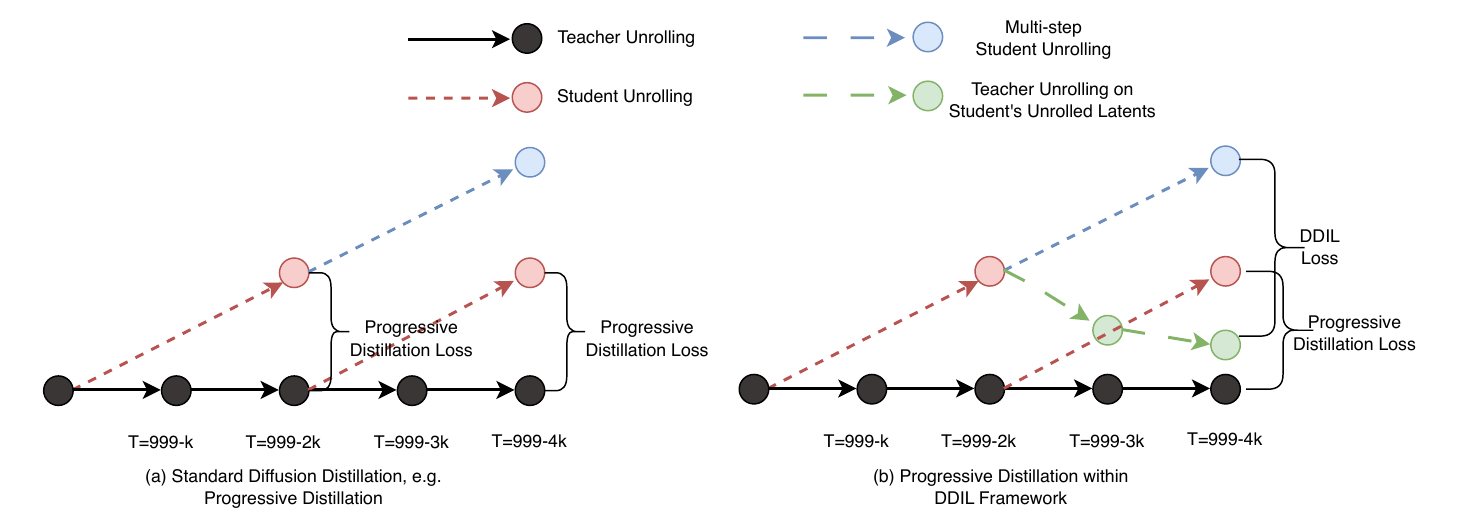}
    \caption{ \textbf{Preditions at different timesteps for different distillation frameworks:} \emph{(a)} We demonstrate standard progressive distillation training framework where student always sees forward diffused latent. \emph{(b)} We show unrolling within our framework which in addition to (a) also obtains distillation feedback by querying teacher (green) on backward trajectory.}
    \label{fig:rollouts}
    \vspace{-15pt}
\end{figure*}

\begin{itemize}\setlength{\itemsep}{-0.2em}

    \item We propose a novel \textit{DDIL} framework which enhances training distribution of the diffusion distillation within the dataset aggregation `DAgger' framework by performing distillation on both the data distribution (forward) and student induced distribution (backward trajectory at inference time), yielding improved aggregate predictive distribution.


    \item  To enhance the stability of the distillation process in diffusion models, we adopt thresholding for both the teacher and student diffusion models to enforce the support of the data distribution with reflected diffusion \cite{lou2023reflected} for distillation. Consequently, this approach further mitigates covariate shift, leading to more substantial improvements when combined with DDIL


    \item We demonstrate that DDIL can be integrated with various distillation approached, and DDIL consistently improves on both quality and diversity within computationally efficient framework. DDIL only requires maintaining a prioritized replay buffer and hence does not incur significant computational or memory overhead.

\end{itemize}

%% file: sections/relatedwork.tex
\section{Related Work}



\noindent \textbf{Diffusion distillation methods.} Progressive distillation \cite{salimans2021progressive, meng2023distillation} and many follow up works \cite{li2023snapfusion,berthelot2023tract} try to reduce the number of iterations of student model by forcing student to mimic multiple steps of the teacher. Consistency models \cite{song2023consistency, luo2023latent, ren2024hyper} assume deterministic probabilistic flow at inference and enforce consistency in the data space for step-distillation.
Additionally, recent work decomposes the diffusion trajectory into multiple segments like in progressive distillation and performs distillation within consistency formulation \cite{kim2023consistency}. Instead of using the real data, methods such as BOOT \cite{gu2023boot} consider bootstrapping in the student trajectory to generate samples of high quality and diversity.
\cite{liu2023instaflow} approximates the underlying map of the pretrained diffusion model as linear paths. While above trajectory level distillation techniques like progressive distillation and consistency-based approaches improve efficiency, the quality of the generated samples exhibits low visual fidelity. 

Alternatively, diffusion distillation has been formulated in the distributional matching framework \cite{yin2023one, luo2024diff, yin2024improveddistributionmatchingdistillation, salimans2024multistep, sauer2023adversarial, sauer2024fast}.
Within distribution matching approaches instead of matching teacher for each trajectory or particle like in previous class of methods, we try to match marginals of distilled student model and pretrained diffusion model.
Further, adversarial loss has been applied to distillation approaches to improve the visual quality of the generated images \cite{sauer2023adversarial,sauer2024fast, lin2024sdxl}.
Most of distributional matching objectives like \cite{sauer2024fast, yin2023one} are mode-seeking and looses on diversity. EM distillation \cite{xie2024distillation} addresses this by richer sampling with langevian MCMC to provide better target for distillation. 

\noindent \textbf{Reverse diffusion as Markov decision process.}
Policy gradient methods have recently gained traction in text-to-image generation with diffusion models by formulating the reverse diffusion process as a markov decision process (MDP) \cite{fan2024reinforcement,XuLWTLDTD23}.
Recent work \cite{fan2024reinforcement} proposes a policy gradient method for data distribution matching in diffusion models. 
\cite{black2023training} introduces a policy gradient algorithm  with reward function that optimizes a diffusion model for downstream tasks.
\cite{yang2024dense} assumes a latent reward function of the reverse denoising process by emphasing the text and image alignment on the coarser steps of image generation. All these approaches have been applied to improve the alignment between the prompts and generated images for high-fidelity synthesis. 
In our work, we leverage the formulation of reverse process as MDP for step-distillation.
This formulation allows interactively update the student model with the observations of the teacher model using dataset aggregation \cite{ross2011reduction}.




%% file: sections/background.tex

\section{Background}\label{sec:background}

\subsection{Reverse Denoising Process as  MDP}
In imitation learning, an agent learns to perform tasks by observing and mimicking the behavior of the expert.
An MDP in imitation learning models the next action based on the previous action and the current knowledge of the environment \citep{ke2021imitation,spencer2021feedback}.
In general, an MDP is represented as $\langle \mathcal{S}, \mathcal{A}, P, \rho_0\rangle$, where $\mathcal{S}$ is a finite set of states, $\mathcal{A}$ is the set of actions, $P(s'| s,a)$ is a state transition kernel to transition from $s$ to $s'$ under the action $a$ and $\rho_0$ is the set of initial states. 
An MDP produces a trajectory which is a sequence of state-action pairs $\tau = (s_0,a_0,s_1,a_1,...,a_T,s_T)$ over $T$ time steps.

We formalize the reverse process of the diffusion models as a finite horizon MDP \citep{black2023training, fan2024reinforcement} with the policy $\pi_{\theta}$ (the diffusion model with parameters $\theta$) where the states and the actions are  $s_t \coloneqq (\mathbf{x}_t,t)$ and $ a_t \coloneqq\mathbf{x}_{t-1}$ respectively.
The transition dynamics is defined by $P(s_{t+1}|s_t,a_t)\coloneqq \delta(\mathbf{x}_{t-1},t)$ and $\rho_0(s)\coloneqq 
(\mathcal{N}(\mathbf{0},\mathbf{I}),T)$ denotes the initial state distribution.  
The trajectory $\tau$ becomes $( \mathbf{x}_T, \mathbf{x}_{T-1}, \ldots, \mathbf{x}_0)$.

\subsection{Co-variate Shift in Diffusion Models} 

With in iterative denoising steps of generation within backward trajectory of diffusion models, student's current predictions determines what the student (learner) sees in next step within sequential setting, which is classic feedback loop \cite{spencer2021feedback} in imitation learning. So if student makes any mistake or has bad score estimate in one of early steps this discrepancy exacerbates in later iterations and results in accumulation of error. This error results in change in input distribution (covariate shift) of latents between training time (forward diffusion) and latents student model encounters when it is unrolled in iterative fashion at generation i.e., backward trajectory. Exposure bias is another closely related line of work \cite{li2023alleviating} which also discusses change in input distribution w.r.t pretrained diffusion model and  propose training-free methods to improve it. Our work primarily focuses on distilling diffusion models and how this shift effects distillation.

Covariate shift is more pronounced for distilled student diffusion model compared to pretrained diffusion model. To further clarify why covariate shift poses more of a challenge for the student model compared to the teacher model, we can consider inference as ancestral sampling (or annealing in score estimation). During generation i.e., within intermediate time-steps of backward trajectory of diffusion model, there is an implicit assumption that the marginal distributions between two consecutive denoising steps significantly overlap, which enables diffusion models to self-correct towards high likelihood. While overlap in marginal distributions of consecutive time-steps is a reasonable assumption in continuous time diffusion models or when the number of denoising steps are sufficiently high,  when considering a diffusion model with only few steps, this assumption does not hold and any we might be querying student  model in out-of-domain at inference time causing performance degradation.

Consequently, any covariate shift  would be more exacerbated for the student model, unlike the continuous time teacher model. We do empirical analysis of covariate shift in \ref{covariate_shift}

\textbf{DAgger to mitigate Co-variate shift:} 
Imitation learning has long been used to learn offline sequential tasks wherein a student model is trained from expert or teacher demonstrations. Standard imitation learning also suffers from covariate shift, i.e., discrepancy in states visited by student and the teacher. Interactive methods such as DAgger \citep{ross2011reduction} in Imitation learning augment training data by querying the teacher model on student's states, thereby obtaining teacher's corrective feedback on states that student has not seen in demonstration data.  Building on the ideas of interactive methods in imitation learning, in our work we aim to improve training distribution for diffusion distillation.

\label{f_div_lfd}\textbf{Imitation Learning as Distribution Matching:} Notably, 
\cite{ke2021imitation} has shown that the imitation learning algorithms can be formalized as $f$-divergence minimization algorithms where the DAgger approach minimizes the total variation (TV) distance between the student and the teacher trajectory distributions. If $\rho_{\theta}(\tau)$ and $\rho_{\eta}(\tau)$ are teacher and student
trajectory distributions respectively, then DAgger minimizes upper bound on total variation. DAgger achieves $O(T\epsilon)$ error compared to behavior cloning equivalent progressive distillation or any method with teacher forcing bounded by $O(T\epsilon^2)$ error.


\subsection{Backward Trajectories for Distilling Diffusion Models} 
Backward trajectory distillation introduced in recent concurrent works like ImagineFlash\citep{kohler2024imagine} and DMD2 \citep{yin2024improveddistributionmatchingdistillation}, focuses solely on evaluating the quality of generated samples without considering the data distribution. Consequently, they lack a mechanism to prevent mode collapse and ensure diversity. 

\textbf{Mode Seeking:} The problem of covariate shift and distribution changes in diffusion distillation is multifaceted. It's not just about input distribution shifts caused by error accumulation. The reduction of diversity in the intermediate steps of backward trajectories (generative process) also plays a crucial role. If diversity is lost early on, it cascades through subsequent steps, limiting the range of possible outcomes. This is akin to error accumulation, but instead of errors, we are consistently losing diversity across time. We can think of it like sequential Monte Carlo sampling in diffusion models: at each step, we are discarding a large number of potential paths (particles), leading to a narrower range of possibilities in the later stages.

While EM Distillation \citep{xie2024distillation} addresses this by employing Langevin MCMC for a richer reverse process and mode-covering divergences, it still doesn't explicitly incorporate the data distribution into its sampling prior during distillation. Depending on right choice of divergence measure, we can consider better mode-preserving distillation objectives and can be integrated with DDIL.

Table \ref{tab:distill_review} provides a summarized overview of the design choices adopted by different techniques. 
\begin{table*}[t!]
\centering
\scriptsize
\caption{Properties of Different Diffusion Distillation Techniques
}
\begin{tabularx}{\textwidth}{@{}Xccccc@{}} 
\toprule
   Model  & $x \sim p_{data}(x)$ &  $x \sim q_{\eta}(x)$  & Preserve  Diversity  \\ 
\midrule    
    Progressive Distillation \citep{meng2023distillation,salimans2021progressive} & \greencheck & \redx  & \greencheck  \\
    ImagineFlash \citep{kohler2024imagine} & \redx & \greencheck  & \redx   \\
    LCM            & \greencheck & \redx  & \greencheck    \\
    InstantFlow \citep{liu2023instaflow}  & \greencheck & \redx   & \greencheck    \\
    ADD \citep{sauer2023adversarial}  & \greencheck & \greencheck  & \redx   \\
    DMD \citep{yin2023one,yin2024improveddistributionmatchingdistillation} & \redx & \greencheck  & \redx \\
    DDIL (Ours) & \greencheck & \greencheck  & \greencheck  \\
    
\bottomrule
\end{tabularx}

\label{tab:distill_review}
\vspace{-0.7em}
\end{table*}

%% file: sections/method_ddil.tex
\section{Method}\label{sec:method}

\subsection{Improving Training Distribution with DDIL}
We introduce Diffusion Distillation with Imitation Learning (DDIL), a novel framework inspired by the DAgger algorithm from imitation learning to enhance the sampling distribution of intermediate noisy latents for distilling diffusion models. Diffusion distillation involves two key considerations: (1) the training distribution of latent states encountered by the student model, and (2) the feedback mechanism employed during distillation. DDIL specifically focuses on improving the training distribution, remaining agnostic to the specific feedback mechanism utilized by different distillation techniques.

To achieve this, DDIL strategically samples intermediate latent variables from three sources: (1) forward diffusion of the dataset, captured by the sampling prior $\beta_{frwd}$ (as illustrated in Algorithm \ref{alg:DDIL_alg}); (2) backward trajectories (unrolled latents) from the student model, denoted by the sampling prior $\beta_{student\_bckwrd}$; and (3) backward trajectories from the teacher model, denoted by the sampling prior $\beta_{teacher\_bckwrd}$, which is particularly advantageous in data-free settings as a proxy in preserving marginal data distribution. Combining these sampling strategies leads to improved distillation performance. 

DDIL is a unified training framework for distilling diffusion models w.r.t sampling prior for distillation. 
DDIL incorporates \textbf{corrective feedback} using $\beta_{student\_bckwrd}$ by preventing accumulation of error on student induced $z_t$ w.r.t distilled model.
Specifically DDIL modifies progressive distillation (PD) and latent consistency models (LCM) by unrolling student and obtain (teacher's) corrective feedback on student's backward trajectories i.e., if a student observes less encountered $z_t^s$ then obtaining a feedback on this rare $z_t^s$ improves student's estimate at $z_t^s$ but in teacher forced distillation methods like PD we only $z_t^s$ at inference and never at training.
 
 Furthermore, while methods like  \cite{kohler2024imagine, yin2024improveddistributionmatchingdistillation} perform distillation on backward trajectories to mitigate covariate shift, they don't account for marginal data distribution during distillation and hence could loose diversity. DDIL addresses this by consistently incorporating feedback from the chosen distillation algorithm on both forward and backward trajectories, i.e., $\beta_{frwd}$ or $\beta_{teacher\_bckwrd}$ are necessary during distillation to better be aware of data distribution which is not the case in methods like GANs,\cite{yin2024improveddistributionmatchingdistillation}, etc.
Our flexible framework thus allows for improved training distribution to boost the performance of diffusion distillation methods.

Algorithm \ref{alg:DDIL_alg} outlines a generalized framework for Diffusion Distillation with Imitation Learning (DDIL). This framework leverages a pre-trained diffusion model (teacher) and a student diffusion model, typically initialized with the teacher's parameters. Additionally, access to real data or equivalent  prompts are assumed, providing representative samples from the marginal data distribution during the distillation process. The framework necessitates specifying hyper-parameters for both the teacher and student models, including their respective discretization schemes. For simplicity we assume DDIM solver in \ref{alg:DDIL_alg}. 
Distillation proceeds by randomly selecting one of three methods for sampling intermediate noisy latent `inputs' to the student model. This selection is governed by user-defined sampling priors: $\beta_{frwd}$, $\beta_{teach\_bckwrd}$, and $\beta_{student\_bckwrd}$, which correspond to the three sources of intermediate latents previously discussed. The choice and updating of these sampling priors, denoted as $\beta_i$, can be tailored based on the training stage, objective function, and overall task goals. If distillation is performed without image data i.e., data-free settings, then $\beta_{frwd} = 0$ and $\beta_{teach\_bckwrd}$ acts as a proxy to sample from data distribution.




Let $q_{\eta}(x)$ be aggregate predictive distribution of distilled student model from its generated trajectories. DDIL objective is to train on sampled latents from student's predictive distribution $q_{\eta}(x)$ and data distribution $p_{data}(x)$
\begin{equation}
    \label{eq:ddil}
    L_{\text{DDIL}} = \mathbb{E}_{t,\epsilon, \tilde{\mathbf{x}} \sim p_{data}(\mathbf{x})} L_\text{{Distill}} + \mathbb{E}_{t,\epsilon, \tilde{\mathbf{x}} \sim q_{\eta}(\mathbf{x})} L_\text{{Distill}}
\end{equation}

Where $L_{\text{Distill}}$ can assume any objective based on chosen algorithm like progressive distillation, latent consistency distillation and distribution matching objective. This makes student model to match it's $q_{\eta}(x)$ to $p_{data}(x)$ not just at $t=0$ but also at other intermediate time-steps and corresponding noisy marginal distributions/

\begin{algorithm}[t!]
\small
\caption{Generalized DDIL framework for Distilling Diffusion Models}\label{alg:DDIL_alg}
\begin{algorithmic}
\Require Teacher diffusion model with text-conditioning with params: $\theta$; student parameters: $\eta$; Dataset $\mathcal{D}$; Time step Discretization  $N,N_s$ of Teacher and student Models respectively.


\State $k = 1000/N$ \Comment{step size teacher diffusion model} 
\State $k_s = 1000/N_s$ \Comment{step size of student diffusion model} 

\State $x \sim D$ \Comment{Sample from data}
\State $T_s \sim \{1000,999,\ldots,1\}$ \Comment{Sample time-step}
\State $\epsilon \sim \mathcal{N}(\mathbf{0},\mathbf{I})$ \Comment{Sample noise}

\Comment{Choose current mini-batch sampling mode $\sim$ [forward, teacher backward, student backward]}

\If {$p \sim U[0,1] < \beta_{frwd} $} \Comment{Forward Process}
    \State $z_t = \alpha_t x + \sigma_t \epsilon$ \Comment{add noise to data}
    \State $z_{T_s} \leftarrow z_t$
\ElsIf {$\beta_{frwd} \leq p < \beta_{teach\_bckwrd} $} \Comment{Teacher Backward}

    \For{$t = \{1000,1000-k,...,T_s\}$} 
        \State $\mathbf{z}_{t-k} = \alpha_{t-k}(\alpha_t \mathbf{z}_t - \sigma_t \hat{\mathbf{v}}_t) + \sigma_{t-k} (\sigma_t \mathbf{z}_t -\alpha_t\hat{\mathbf{v}}_t)$
        \State $t \leftarrow t-k$
    \EndFor
    \State $z_{T_s} \leftarrow z_{t-k}$

\Else \Comment{Student Backward $\beta_{student\_bckwrd}$}
    \For{$t = \{1000,1000-k_s,...,T_s\}$} 
        \State $\mathbf{z}_{t-k_s} = \alpha_{t-k_s}(\alpha_t \mathbf{z}_t - \sigma_t \hat{\mathbf{v}}_t^s) + \sigma_{t-k_s} (\sigma_t \mathbf{z}_t - \alpha_t \hat{\mathbf{v}^s}_t)$
        \State $t \leftarrow t-k_s$
    \EndFor
    \State $z_{T_s} \leftarrow z_{t-k_s}$

\EndIf
\State Train student diffusion model on $z_{T_s}$ with distillation method.


\end{algorithmic}
\vspace{-4pt}
\end{algorithm}

\textbf{Reflected Diffusion Distillation:}
When distilling diffusion models either the teacher or student model might not satisfy implicit assumed support during distillation which could makes training unstable and require large batch sizes, etc. We adopt reflected diffusion models \cite{lou2023reflected} framework for distillation i.e., threshold 
score estimates of teacher model and/or student model. 

Static thresholding \cite{saharia2022photorealistic} is applied to the teacher model's estimates consistently across all investigated methods: progressive distillation, Latent Consistency Models (LCM), and DMD2. Furthermore, within the consistency distillation framework, thresholding is also applied to the target derived from the student model. In case of DMD2, thresholding is applied to the score estimates of the pre-trained diffusion model, the fake critic, and the student model. Without thresholding, our gradient feedback could be noisy and  negatively impacting training stability. We observe that when training DMD2 with small batch (7), without thresholding performance deteriorates and when we adopt thresholding can observe monotonic improvement in performance.

\subsection{DDIL Integration}
This section examines the integration of DDIL with various distillation techniques. Detailed design choices are further elaborated in the appendix (section to be updated).

\textbf{PD + DDIL:} DDIL is integrated with progressive distillation using a DAgger-inspired approach \cite{ross2011reduction}. Distillation is performed on mixed rollouts generated by alternating between the pre-trained and student diffusion models within each generation. A stateless DDIM solver facilitates this interleaved sampling process.

\textbf{Computational overhead:} DDIL only requires maintaining a replay Buffer $D_{Dagger}$  which stores 
intermediate latents from interactive reverse process i.e.,  switching between teacher and student models using $\hat{x}_t$ (Algorithm 1 of supplementary). The buffer is updated periodically by replacing older data with new data and only needs to store the input latents for training. Empirically, when applying DDIL to PD (method with the most gradient updates), we observed less than a 5\% increase in training time.

\textbf{LCM + DDIL:} DDIL is also applied to consistency distillation. Due to the pre-trained model's lack of prior consistency training, the mixed-rollout strategy used in progressive distillation is not directly applicable. Therefore, DDIL is extended to LCM by applying consistency distillation to both forward and backward trajectories of the student model, leveraging the student-induced distribution and demonstrating performance improvements.

\textbf{DMD2 + DDIL:} Mirroring the progressive distillation approach, mixed rollouts are employed within the DMD2 framework. Trajectories are sampled up to a predetermined noise level or timestep (e.g., t=500) using either the student or teacher model. The resulting latent serves as input to the student model for gradient feedback within the DMD2 formulation.
Distribution Matching aligns well within our DDIL framework as discussed in \Cref{f_div_lfd} where within imitation learning framework we consider matching generated trajectory distributions of student model with teacher or expert's trajectory distribution (of states or equivalently noisy latents in case of diffusion).

DDIL integration does not incur significant additional compute as we collect sampled intermediate time-step latents into a dataset and updating this dataset at a significantly lower frequency compared to gradient feedback, exactly like DAgger and elaborated in context of progressive distillation in supplementary material.

%% file: sections/experiments.tex
\section{Experiments}
\label{sec: experiments}


\textbf{Datasets and metrics:} 
Following standard practice for evaluating text-to-image diffusion models \cite{rombach2022high, meng2023distillation}, we evaluate our distilled models zero-shot on two public benchmarks: COCO 2017 (5K captions), and COCO 2014~\cite{lin2014microsoft} (30K captions) validation sets.  
We use each caption to generate an image with a randomized seed and report CLIP score using OpenCLIP ViT-g/14 model \cite{ilharco2021openclip} to evaluate image-text alignment. We also report Fréchet Inception Distance (FID) \cite{heusel2017fid} to estimate perceptual quality. 

To measure fidelity and diversity of generations we report Density-Coverage, Precision-Recall \cite{stein2024exposing, kynkaanniemi2019improved, naeem2020reliable} and also $LPIPS_{Diversity}$. 
Coverage improves upon the recall metric to better quantify diversity by building nearest neighbour manifolds around the real samples, instead of the fake samples, as they have less outliers.
We use a more effective and robust DINOv2 \cite{oquab2023dinov2} feature space to on 30k prompts of COCO2014 to report Density-Coverage, Precision-Recall metrics across all experiments.
In case of $LPIPS_{Diversity}$, for a given prompt we generate output for 10 different seeds and obtain pair-wise LPIPS score and finally average over 50 randomly sampled COCO 2017 prompts.

\textbf{Training:}
For all our experiments, we choose AdamW optimizer \cite{loshchilov2017decoupled} with $1e-05$ learning rate with warmup and linear schedule on a batch size of 224 in case of progressive distillation, 360 in case of LCM and 7 in case of DMD2 on SSD1B. To optimize for GPU usage, we adopt gradient checkpoint and mixed-precision training. Please refer to Appendix A  for additional training details.

\textbf{PD + DDIL:}In case of progressive distillation, we train the model for 4k steps for $\epsilon$ to $v$ space conversion to perform step distillation in $v$ space \cite{salimans2021progressive}. Then we perform guidance conditioning following the same protocol as \cite{meng2023distillation} where we sample guidance scale $\omega \sim [2,14]$ and incorporate additional guidance embedding as in \cite{rombach2022high} followed by step distillation. Overall we train $~10K$ steps to obtain guidance conditioned checkpoint SD$(gc)$. For progressive distillation, we start with a 32-step discretization assumption for the pre-trained diffusion model and perform $32 \to 16$ step distillation for 5K iterations with 500 steps of warm-up. We progressively increase training compute or gradient steps as we go towards fewer iteration student. We share more details in Section 2 of Appendix.


\textbf{LCM + DDIL:} We trained both LCM and DDIL models on the Common Caption dataset for 8,000 steps, using the SDv1.5 checkpoint and a batch size of 60 on 6 A100 GPUs. To enhance consistency distillation, we introduced backward trajectory sampling. Specifically, we randomly select a number of inference steps (3, 4, or 5) and obtained samples at specific timesteps along the backward trajectory. This enabled us to incorporate consistency distillation loss feedback not only on forward diffused latents but also on these backward trajectory latents within our framework.

\textbf{DMD2 + DDIL:}
In this work we consider distilling SSD1B checkpoint \cite{gupta2024progressive} with DMD2 for computational efficiency. 
To achieve stable training within the DMD2 framework, which utilizes a teacher model and a `fake' critic, we update the fake critic ten times for every update of the student model as we adopt `SSD1B' as critic instead of SDXL and hence allow more updates.

We distill SSD1B with DND2 and also integrate DDIL trained with significantly small batch size of 7 and 40,000 gradient steps on a single A100 node. 
By adopting reflected diffusion distillation, we achieve improved training stability and a significant boost in performance, both quantitatively and qualitatively.
Further performance gains are observed when incorporating a mixed rollout setting of DDIL within DMD2, as demonstrated in the table \ref{tab:sota_t2i_dmd2}. We can observe consistent boost in both quality i.e., FID@30k of \textbf{19.41\%} and diversity i.e, coverage boost of \textbf{19.38\%} over baseline DMD2 when DDIL is integrated.

\textbf{Student Selection Prior: }Our protocol for student selection in trajectory collection follows standard practice from imitation learning. Where early in training, student's performance is bad and hence we prioritize sampling more from $p_{data}(x)$ but as training progress and student's performance is good we want to obtain expert feedback on mistakes that student makes i.e., address co-variate shift caused by feedback and training-inference mismatch but still sample from $p_{data}(x)$ to preserve marginal data distribution.

\begin{table*}[t!]
\scriptsize
\centering
\caption{Text guided image generation results on \textbf{MS-COCO 2014-30K} validation set except for CLIP Score which is evaluated on \textit{COCO 2017-5K}. Our 4-step Model demonstrates SOTA performance for checkpoints based on SD1.5 and improves on diversity}
\begin{tabularx}{\textwidth}{@{}Xccccccccc@{}}
\toprule
\textbf{Model} & \textbf{Steps} & \textbf{NFEs} & \textbf{FID@5K} \([\downarrow]\) & \textbf{FID@30K} \([\downarrow]\) & \textbf{CLIP} \([\uparrow]\)  & \textbf{Precision}  \([\uparrow]\) & \textbf{Recall}  \([\uparrow]\) & \textbf{Density}  \([\uparrow]\) & \textbf{Coverage}  \([\uparrow]\) \\

\midrule
Progressive Distillation & 4 & 4 & 23.34& 14.72 & 0.302 & 0.595 & 0.605 & 0.301 & 0.397 \\
\hspace{0.5cm} + DDIL & 4 & 4 & \textbf{22.42}& 13.97 & 0.302 & 0.603 & 0.605 & 0.318 & 0.411 \\
Progressive Distillation & 2 & 2 & 26.43 & 16.46 & 0.288 & 0.392 & 0.543 & 0.153 & 0.202 \\
\hspace{0.5cm} + DDIL & 2 & 2 & 24.13 & 15.81 & 0.291 & 0.455 & 0.547 & 0.189 & 0.249 \\

\midrule
LCM & 4 & 4  & 24.39  & 16.45 & 0.305  & 0.668 & 0.687 & 0.376 & 0.481 \\ 
\hspace{0.5cm} + Reflected & 4 & 4 & 24.25  & 16.44 & 0.306  & 0.670 & 0.687 & 0.378 & 0.478 \\
\hspace{0.5cm} + DDIL & 4  & 4 & 23.44 & 15.87 & 0.308 &  0.672 & 0.690 & 0.382 & 0.493 \\ 
\hspace{0.5cm} + Reflected + DDIL  & 4 & 4  & 22.86 & 15.34 & 0.309  & 0.669 & 0.692 & 0.381 & 0.499 \\
\hspace{0.5cm} Improvement ($\mathbf{\Delta}$ \%) & 4 & 4  & \bfseries 6.3\% & \bfseries 6.7\% & 1.31\%  & 0.14\% & 0.72\% & 1.32\% & \bfseries 3.74\% \\

\midrule
SD [\(v\)] & 32 & 64 & 22.50 & 13.48 & 0.321 & - & - & - & - \\
SD [\(gc\), Teacher] & 32 & 32 & 24.46 & -& 0.304 & - & - & - & - \\

\bottomrule
\end{tabularx}   
\label{tab:sota_t2i_coco2017}
\end{table*}

\subsection{Text-guided Image Generation}
We demonstrate effectiveness of our proposed DDIL framework across different baseline distillation techniques in case of text-to-image generation tasks as shown in \Cref{tab:sota_t2i_coco2017}. DDIL consistency improves on progressive distillation(PD) and latent consistency models (LCM) as observed in \Cref{tab:sota_t2i_coco2017}/ In case of progressive distillation, for 4-step version DDIL improves FID from $23.34 \to 22.42$ and maintains clip score of $0.302$ and similarly we can also observe DDIL improves on LCM with FID from $24.25 \to 22.86$ and CLIP score $0.306 \to 0.309$. From Tab. 4 in appendix, we can observe that
4-step variant of $PD+DDIL$ with a guidance value of 8 achieves best FID of 13.97, the highest among trajectory based distillation methods. 

We also demonstrate effectiveness of DDIL with distribution matching techniques which adopt multi-step student like in DMD2. When we apply DMD2 to SSD1B, we can observe that DDIL improves FID from $31.77 \to 27.72$ i.e., 12.7\% improvement and clip score from $0.320 \to 0.326$ and HPSv2 score from $0.302 \to 0.304$.

\textbf{Computational efficiency:} 
DDIL demonstrates superior computational efficiency compared to state-of-the-art methods such as Instaflow and DMD. For instance, LCM augmented with DDIL (LCM+DDIL) achieves strong performance using only 8,000 gradient steps with a batch size of 420. This contrasts sharply with Instaflow, which requires 183 A100 GPU-days for distillation. DDIL with progressive distillation (PD) reduces this to 15 A100 GPU-days. Similarly, while DMD necessitates 64 GPUs with a larger batch size and extended training duration, DDIL attains comparable results using significantly fewer resources.

Compared to 64 A100s with batchsize of 128 reported in DMD2, in our work we only use 7 A100 GPUs with batchsize of 7 and demonstrate comparable performance with DMD2 on SSD1B, even with worse critic especially when we adopt reflected diffusion to improve stability of training and then incorporating DDIL further improves performance. DDIL demonstrates strong performance with DMD2, LCM using significantly smaller batch sizes and fewer gradient steps demonstrating its generalization across methods and potential for wider appplication when distilling diffusion models.



\textbf{Diversity vs. Quality trade-off:} 
Diffusion distillation aims to compress a pre-trained diffusion model while maintaining performance. This compression is achieved by reducing compute i.e., iterations. However, this presents a fundamental challenge: achieving both high quality and diverse generations with significantly reduced model capacity. Within distillation our goal is to mimic trajectory or map of pretrained diffusion model in significantly fewer steps and hence this could result in discretization errors which compound across steps resulting in poor performance.


So depending on distillation method we choose fidelity or diversity, adversarial distillation methods \cite{sauer2023adversarial, sauer2024fast}  exhibit a decrease in generation diversity compared to the baseline but demonstrate good quality. As discussed in previous section most common objectives in distribution matching adopt mode-seeking objectives \cite{xie2024distillation,yin2024improveddistributionmatchingdistillation} where as methods like progressive distillation preserve map from prior to data-distribution and hence preserves diversity but looses on quality because of discretization errors, where as other methods can ignore high curvature or difficult regions of initial map there by choosing quality over diversity. There is need for future work to improve diversity, as both DMD2 and LCM have poor coverage, recall on SSD1B.
\begin{table*}[t!]
\centering
\scriptsize
\caption{\small{Text guided image generation results on \textbf{COCO 2014-30K} validation set except for CLIP Score which is evaluated on \textit{COCO 2017-5K} . These results are obtained by adopting latent consistency distillation retrained for \textbf{SSD1B} and incorporating DDIL within DMD2 setting. We integrate DDIL into the DMD2 framework by unrolling just the student model exactly like DMD2 but also unrolling teacher to corresponding noise level too to better capture underlying data distribution and align gradient fields of student model and teacher model.We use Guidance = 8 for teacher, LCM\(*\) checkpoint was not trained by us.}}
\begin{tabularx}{\textwidth}{@{}Xccccccccc@{}}
\toprule
\textbf{Model} & \textbf{Steps} & \textbf{FID@5K} \([\downarrow]\) & \textbf{FID@30K} \([\downarrow]\)  & \textbf{CLIP} \([\uparrow]\)  & \textbf{HPSV2} \([\uparrow]\)  & \textbf{Precision} \([\uparrow]\) & \textbf{Recall} \([\uparrow]\)  & \textbf{Density} \([\uparrow]\) & \textbf{Coverage} \([\uparrow]\) \\
\midrule
SSD1B (Teacher) & 20  & 30.23 & 20.71  & 0.336 & 0.297 & 0.463 & 0.679 & 0.186 & 0.271\\

SSD1B-LCM\(*\) & 4  & 35.23 & 28.46 & 0.311 & 0.282  & 0.336 & 0.586 & 0.117 & 0.171 \\


\midrule
SSD1B-DMD2 & 4 & 31.77 & 23.24 & 0.320 & 0.302 & 0.361 & 0.605 & 0.136 & 0.196 \\


\hspace{0.5cm} +DDIL & 4   & 27.72   &  18.73 &  0.326 &  0.304 & 0.426 & 0.658 & 0.163 & 0.234 \\ 
\hspace{0.5cm} Improvement ($\mathbf{\Delta}$ \%) & 4  & 12.74\%  & \bfseries 19.41\% & 1.88\% & 0.66\% & \bfseries 18.05\% & \bfseries 8.76\% & \bfseries 19.85\% & \bfseries19.38\% \\ 

\bottomrule
\end{tabularx}
\label{tab:sota_t2i_dmd2}
\end{table*}





\subsection{Covariate shift Analysis}
\label{covariate_shift}


We investigate covariate shift stemming from error accumulation in distilled diffusion models. Our analysis utilizes a 32-step Classifier-Free Guidance (CFG) teacher and a 4-step student trained via progressive distillation (PD) and evaluate on the MS-COCO 2017 (5K) validation set. We conduct mixed-rollout evaluations, leveraging a stateless DDIM solver to stochastically select between the teacher and student at each step (t-250), with a prior probability $p_T$ of choosing the teacher. 

Given that we are working with PD checkpoint, if there is no discrepancy between teacher (black) trajectory and student's trajectory (red+blue) in Fig. \ref{fig:rollouts} we should observe same performance irrespective of relative prior on teacher, i.e., even when we switch to teacher on student's trajectory (green transitions) in Fig. \ref{fig:rollouts} performance should not improve. In Tab. \ref{tab:selection_rate} we control choosing teacher at different intermediate time steps and in absence of covariate shift increasing teacher selection rate (green transitions) should not vary performance but we can see that as we increase prior on teacher we see consistent improvement in CLIP score and FID validating hypothesis on presence of covariate shift.



\begin{table}[t!]
\centering
\scriptsize
\begin{minipage}[t]{.45 \textwidth}
\caption{\small \textbf{Covariate shift Analysis:} Evaluation with different teacher selection rates.} \label{tab:selection_rate}
\begin{tabularx}{\textwidth}{@{}Xcc@{}} 
\toprule

   \textbf{$p_{T}$} & \textbf{FID [$\downarrow$]} & \textbf{CLIP [$\uparrow$]}  \\ 
\midrule
                  0.8   & 23.14  &  0.319 \\
                  0.6   & 22.33  &  0.317 \\
                  0.4   &  21.95 & 0.313  \\
                  0.2   &  21.92 &  0.307 \\
                  0.0 & 22.42 & 0.302 \\
\bottomrule
\end{tabularx}
\end{minipage}
\vspace{-0.5 cm}
\end{table}

%% file: sections/supplementary.tex
\newpage
\appendix

{
\centering
\Large
\textbf{DDIL: Diversity Enhancing Diffusion Distillation with Imitation Learning} \\
\vspace{0.5em}-- Supplemental Material -- \\
\vspace{1.0em}
}



\begin{figure*}[h]
  \centering
  \begin{minipage}{1.0\textwidth}
    \includegraphics[width=\linewidth]{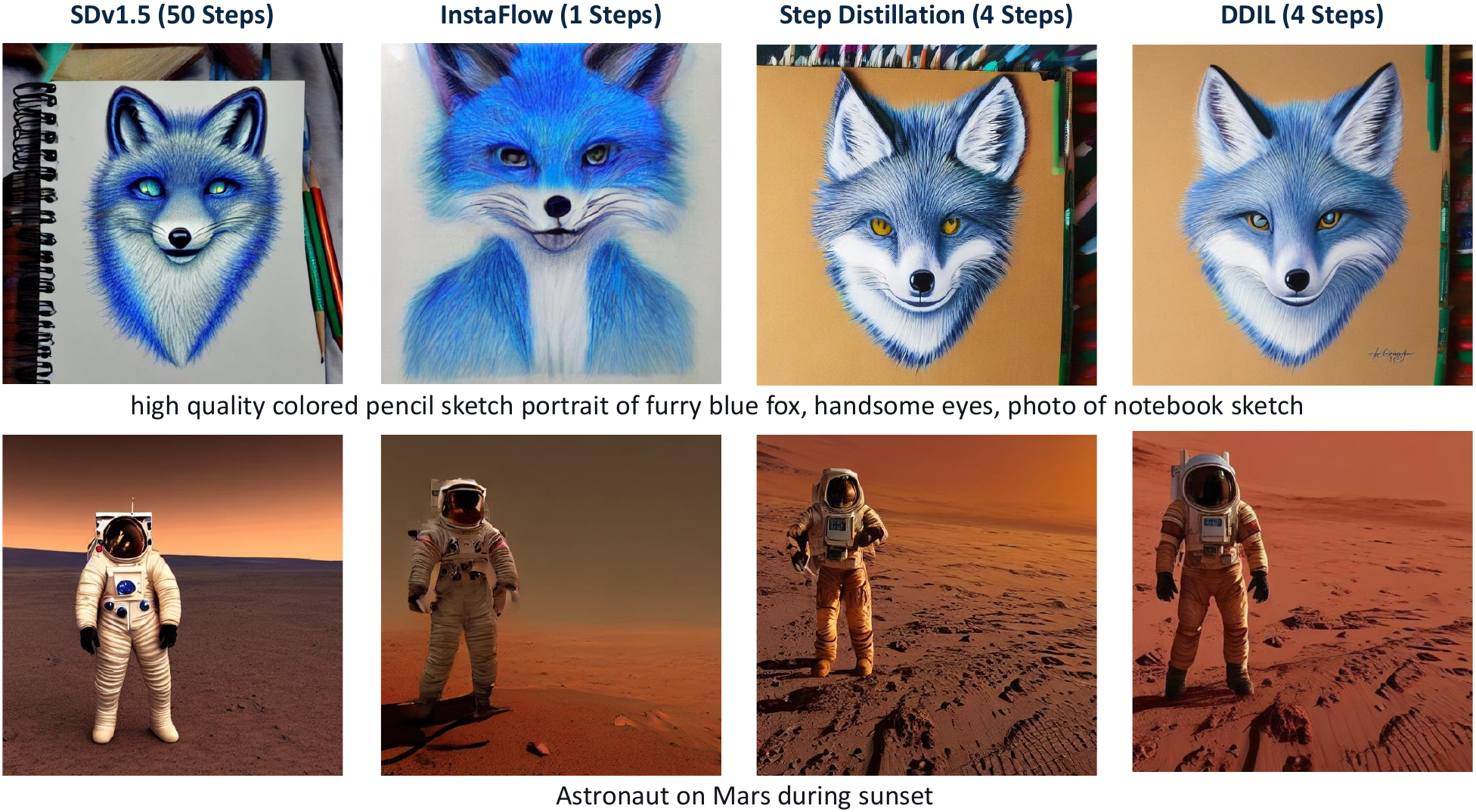}
    \label{fig:image1}
  \end{minipage}
  \begin{minipage}{1.0\textwidth}
    \vspace*{-0.4cm}
    \includegraphics[width=\linewidth]{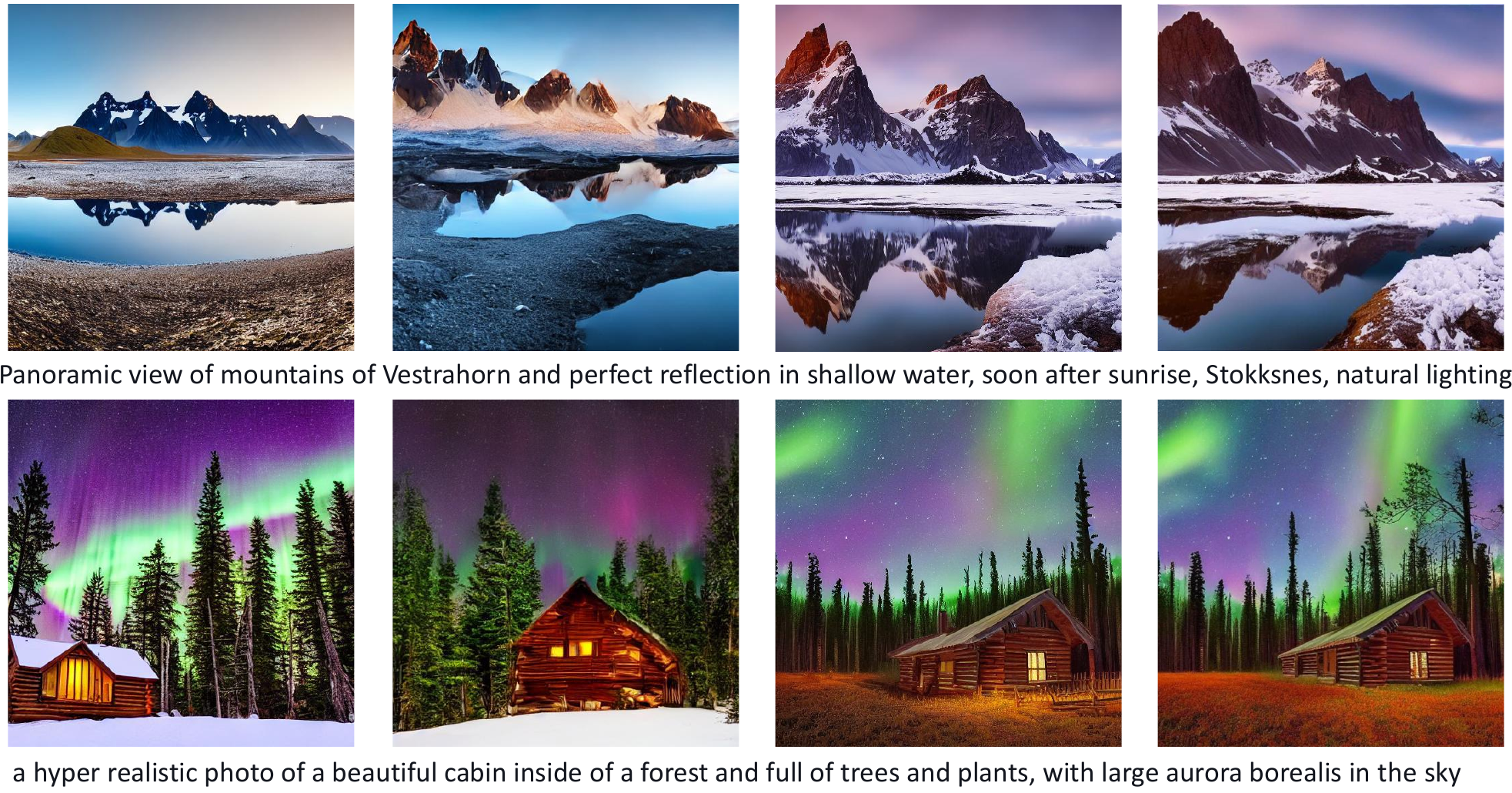}
    \label{fig:image2}
  \end{minipage}
  \vspace{-0.5cm}
  \caption{Qualitative comparison of images generated with different distillation techniques. We can observe that DDIL improves progressive distillation(PD), for e.g., we can observe `astronaut' slightly disfigured in case of PD but DDIL(+PD) quality is good.}
\end{figure*}

\section{ Additional Training Details}
\label{pd_appendix_trn}

In case of progressive distillation (PD), we use DPM++ multi-step \cite{lu2022dpm++} solver (2-step) for all evaluations except co-variate shift analysis, where a state-less solver like DDIM \cite{zhang2022gddim} will enable switching between two different models and corresponding reverse processes. We train our model on an internal text-to-image dataset and only perform distillation without updating the teacher model. In case of consistency distillation with LCM, we adopt the default LCMScheduler. In the case of DMD2, we follow the same settings as the original repository except that in our work, we focus on SSD1B \cite{gupta2024progressive} instead of SDXL to reduce computational requirements for training. By default, we adopt reflected diffusion for all three diffusion distillation techniques we considered in this work.

\paragraph{Distribution Matching Distillation(DMD2):}

In case of DMD2, to integrate DDIL in addition to unrolling student model we also unroll pre-trained or teacher model. This makes student model to be well aligned with pre-trained diffusion model and also better capture underlying statistics. In addition this obtains better feedback within score distillation framework as we not only obtain teacher's feedback w.r.t student's generation at terminal states but w.r.t each of student's denoising step when input to student model is initialized by unrolling pre-trained teacher model until corresponding input timestep.

When integrating DDIL within DMD2, we do not consider forward diffusion and instead unroll teacher model and assume pre-trained diffusion model is a proxy for marginal data distribution. In principle we can consider real data to better capture data distribution i.e., tighter ELBO but we leave it for future work. To integrate DDIL within DMD2, we found sampling priors of 
\textit{$\beta_{teach\_bckwrd} = 0.6,\beta_{student\_bckwrd} = 0.4$} and as we don't consider real data to obtain score distillation feedback $\beta_{frwd} = 0.0$

\paragraph{Latency Consistency Models(LCM):} In case of LCM \cite{luo2023latent}, we explored a setting where we add backward trajectories while distillation in addition to forward diffusion, i.e., we found significant boost even without unrolling teacher. This would be mean more sampling at training compared to interactive denoising considered in case of Progressive Distillation, where we collect an independent dataset exactly following DAgger \cite{ross2011reduction} protocol. We can consider either protocols to incorporate DDIL with distillation methods. In case of LCM 50\% of time we train following standard protocol and remaining 50\% of time, we unroll student model with LCMScheduler to a random step within total number of inference steps $\in {2,3,4,5,6}$ to obtain feedback to correct covariate shift within consistency distillation formulation. We adopt sampling priors of \textit{$\beta_{frwd} = 0.5,\beta_{teach\_bckwrd} = 0.0,\beta_{student\_bckwrd} = 0.5$} in case of LCM.

\paragraph{Progressive Distillation:} 
For $16 \to 8$, we follow a similar protocol as $32 \to 16$, but we further split each stage of training into two parts.
First, we do distillation for 6K steps to obtain a checkpoint and resume with warmup and $1e-05$ learning rate for another 4K steps of training. For $8 \to 4$ and $4 \to 2$ distillation, we first distill model for 8K steps followed by another 6K steps.We split single stage of training into two parts to exactly match training protocol of Step Distillation and DDIL. See supplemental for additional details. We adopt timesteps for discretization from the default config of the DPM++ solver i.e., for 4-step models our timesteps are $\{999,749,500,250\}$. 

We follow similar protocol as progressive distillation for PD+DDIL but as we now distill using on text-to-image data using forward process and interactively mixed unrolled trajectories, we have this additional sampling prior hyper parameters \textbf{$\beta_{frwd},\beta_{teach\_bckwrd},\beta_{student\_bckwrd}$}.

\textit{Sampling Priors for PD:} For $16 \to 8$ distillation $75\%$ we obtain $z_t$ using forward diffusion i.e, $\beta_{frwd}=0.75$ making this part of training exactly equivalent to step Distillation. For remaining $25\%$ we choose data from mixed interaction unrolled trajectories where for first 6K steps, student is only selected $15\%$ of time for $z_{t} \to z_{t-2k}$ transition for trajectory collection, where as for next 4K iterations we choose student $80\%$ of time and train $50\%$ of time on collected trajectories. 

Effectively for first 6K steps of training we have prior $x \sim p_{data}(x)$ as $\beta_{frwd}=0.75$, $\beta_{teach\_bckwrd} = 0.25*0.85 = 0.2125$ and $\beta_{student\_bckwrd} = 0.25*0.15 = 0.0375$. 
And for next 4K iterations, $\beta_{frwd}=0.5$ and  $\beta_{teach\_bckwrd} = 0.5*0.2 = 0.10$ and $\beta_{student\_bckwrd} = 0.5*0.8 = 0.40$.

And follow same sampling prior protocol for two parts of training for each stage of progressive distillation with same hyper-parameters as step distillation(PD). Overall we need $~40K$ updates to obtain a 4-step checkpoint or $~55K$ updates to obtain 2-step checkpoint for PD or PD+DDIL, as our sampling is parallelized across GPUs we observe $<5\%$ overhead for DDIL over step distillation, it takes 2 days on single node of 8 A100 GPUs to perform distillation.

\section{Additional Evaluation Results}

\begin{table*}[t!]
\scriptsize
\centering
\caption{Text guided image generation results on \(512 \times 512\) \textbf{MS-COCO 2017-5K} validation set. Our 4-step Model demonstrates SOTA performance for checkpoints based on SD1.5 whereas ADD is based on SD2.1 with a more expressive text encoder and LCM also uses a different checkpoint. \(*\) denote derived baselines and \(\ddagger\) denote different checkpoints}
\begin{tabularx}{\textwidth}{@{}Xccccc@{}}
\toprule
\textbf{Model} & \textbf{Steps} & \textbf{NFEs} & \textbf{FID} \([\downarrow]\) & \textbf{CLIP} \([\uparrow]\) & \textbf{LPIPS\(_{Diversity}\)}  \([\uparrow]\) \\
\midrule
SnapFusion\(*\) \cite{li2023snapfusion} & 8 & 16 & 24.20 & 0.300 & - \\
Step Distillation\(*\) \cite{meng2023distillation} & 8 & 8 & 26.90 & 0.300 & - \\ 
Step Distillation\(*\) \cite{meng2023distillation} & 4 & 4 & 26.40 & 0.300 & - \\ 
UFOGen\(*\) \cite{xu2023ufogen} & 1 & 1 & 22.5 & \textbf{0.311} & - \\ 
ImagineFlash\(*\) \(\ddagger\) \cite{kohler2024imagine} & 2 & 2 & 34.7 & 0.301 & - \\
\midrule
ADD \(\ddagger\) & 1 & 1 & 19.7 & 0.328 & 0.52 \\
LCM \(\ddagger\) & 4 & 4 & 36.36 & 0.294 & 0.49 \\
LCM-LoRA \(\ddagger\)& 4 & 4 & 37.01 & 0.300 & 0.52 \\
LCM-LoRA & 4 & 4 & 36.46 & 0.291 & 0.61 \\
\midrule
LCM & 4 & 4  & 24.39 & 0.305  & 0.61 \\ 
\hspace{0.5cm} + Reflected & 4 & 4 & 24.25 (-0.6\%) & 0.306 & 0.59 \\
\hspace{0.5cm} + DDIL & 4  & 4 & 23.44(-3.9\%) & 0.308 & 0.59 \\ 
\hspace{0.5cm} + Reflected + DDIL  & 4 & 4  & 22.86 (-6.3\%) & \textbf{0.309} & 0.59 \\

\midrule
Progressive Distillation & 4 & 4 & 23.34 & 0.302 & 0.60 \\
\hspace{0.5cm} + DDIL & 4 & 4 & \textbf{22.42} & 0.302 & 0.60 \\
Progressive Distillation & 2 & 2 & 26.43 & 0.288 & 0.58 \\
\hspace{0.5cm} + DDIL & 2 & 2 & 24.13 (-8.7\%) & 0.291 & 0.58 \\
\midrule
SD (\(v\)) & 32 & 64 & 22.50 & 0.321 & 0.62 \\
SD (\(gc\)) & 32 & 32 & 24.46 & 0.304 & 0.62 \\

\bottomrule
\end{tabularx}   
\label{tab:sota_t2i_coco2017}
\end{table*}

In \Cref{tab:sota_t2i_coco2014} we demonstrate that DDIL achieves best performance among trajectory based distillation methods. 

\begin{table*}[t]
\centering
\scriptsize
\caption{\small{Text guided image generation results on $256\times256$ \textbf{COCO 2014} val set.}}
\begin{tabularx}{\textwidth}{@{}Xccccc@{}}
\toprule
\textbf{Model} & \textbf{Steps}  & \textbf{FID} [$\downarrow$]  \\
\midrule
LCM-LoRA (4-step)  & 4  & 23.62   \\
LCM-LoRA (2-step) & 2  & 24.28   \\
DMD  & 1  & 14.93 \\
\midrule
Progressive Distillation (PD) & 4  & 14.72  \\
\hspace{0.5cm} \textit{+DDIL} & 4 & \textbf{13.97}   \\
Progressive Distillation (PD) & 2  & 16.46 \\
\hspace{0.5cm} \textit{+DDIL} & 2 & 15.81  \\
\midrule
SD & 50  & 13.45 \\
\bottomrule
\end{tabularx}
\label{tab:sota_t2i_coco2014}
\end{table*}

\section{Diffusion Distillation Methods}

\subsection{Progressive Distillation}
Progressive Step distillation aims at reducing the number of timesteps $T$ of the sampling (reverse) process in the diffusion models by learning a new student model.
Starting from timestep $t$ within the reverse diffusion process, given the discretization interval $k$, $N$ steps of the teacher are distilled into $N/k$ steps of the student ($T=N$ in the first iteration of step-distillation) \cite{salimans2021progressive}.
We query the teacher model at timesteps $t-k$ and $t-2k$ while the student estimates are obtained at $t-2k$ using a DDIM \cite{song2020denoising}.

Following the formulation in \cite{li2023snapfusion}, the teacher model is unrolled for two DDIM steps to $t-k$ and $t-2k$ starting at timestep $t \in \left[T\right]$ and $0 \leq t-2k <{t-k}$ with input noisy latent $\mathbf{z}_t$ while student model performs one denoising step.
Where $\hat{\mathbf{v}}_t^{\text{s}}$ is the velocity estimate from the student model $\hat{\mathbf{v}_{\eta}(\mathbf{z}_t,t)}$. The student model predicts the latent $\mathbf{z}_{t-2k}^\text{s}$ from $\mathbf{z}_t$ of the teacher and thus    $\mathbf{z}_{t-2k}^\text{s}=\mathbf{z}_{t-2k}$. Progressive distillation loss with guidance conditioned teacher is denoted by, 
\begin{equation}
    L_\text{{PD}} = \max\left(1,\frac{\alpha_t^2}{\sigma_t^2}\right)\left\| \hat{\mathbf{x}}_t^s - \frac{\mathbf{z}_{t-2k} - \frac{\sigma_{t-2k}}{\sigma_t}\mathbf{z}_t}{\alpha_{t-2k}-\frac{\sigma_{t-2k}}{\sigma_t}\alpha_t}\right\|_2^2.
    \label{eq:step_distil}
\end{equation}
Here, the student model is trained with teacher forcing as is evident in \Cref{eq:step_distil}. 
During sampling from the student model, the teacher observations are not provided and therefore, the student model can drift from the expected trajectory \cite{huszar2015not}.

\subsection{Consistency Models}

While multi-step extensions of consistency distillation decompose the trajectory and enforce consistency within segments they remain susceptible to covariate shift with respect to the backward trajectory. This stems from the inherent discrepancy between the teacher and student model's perception of the data distribution in backward trajectory, which is addressed by our proposed DDIL framework. Hence, benefits of DDIL are complementary and extend to multi-step Consistency Distillation variants like CTM and TCD. 




\section{Progressive Distillation with DDIL}

Inspired by the success of the interactive learning DAgger algorithm in imitation learning and following the formulation of the reverse process of the diffusion models as probability flow ODE, we first extend the DAgger framework to diffusion models 
by considering the higher iteration denoising model as expert and fewer iteration denoising model as a student in \Cref{alg:dagger}
Following this, in \Cref{alg:step_distill}, we present the complete DDIL approach with interactive learning. 

For sampling in diffusion models, the student predicted latent $\mathbf{z}_t$  is aligned with the teacher trajectory by adding the state-action pair $(\mathbf{z_t},\epsilon,t) \in \tau_{\theta}$ to the aggregated dataset.
Note that the dataset aggregation is done randomly so that the model is aware of the teacher and the student's distributions.

In our DDIL algorithm outlined in \Cref{alg:step_distill}, the distillation is performed iteratively by taking the sample from the aggregated dataset 
or from the default training dataset with forward diffusion.
Following this, two steps of DDIM sampling are performed on the teacher model to obtain the estimate $\mathbf{z}_{t-2k}$ and subsequently optimize \Cref{eq:step_distil}. This framework introduces a self-correcting behavior. Even if the student deviates from the teacher's trajectory at any step of the reverse diffusion process -- the teacher can provide corrective feedback.

\begin{algorithm}[t!]
\small
\caption{Interactive Trajectory collection for Dataset Aggregation (DAgger)}\label{alg:dagger}
\label{DAgger for Progressive distillation}
\begin{algorithmic}
\Require Teacher diffusion model with text-conditioning with params: $\theta$; teacher velocity: $\mathbf{v}$; student model velocity: $\mathbf{v}^s$; student parameters: $\eta$
\Require Initialize DAgger dataset to collect trajectories, $ \mathcal{D}_{DAgger}  \leftarrow \emptyset $
\Require Student diffusion models with text-conditioning and parameters $\eta$
\State $\mathbf{x}_T \sim \mathcal{N}(\mathbf{0},\mathbf{I})$
\State $t = T = 1000$
\State $k = 1000/N$ \Comment{step length under assumed discrete setting of current teacher diffusion model} 
\For{$t = \{1000,1000-2k,...,1\}$} 
    \If{$p \sim U[0,1] < \beta $} \Comment{Choosing Student model vs teacher model for current iteration}
        \State $ \# \text{ One step of student DDIM step}$
        \State $\mathbf{z}_{t-2k} = \alpha_{t-2k}(\alpha_t \mathbf{z}_t - \sigma_t \hat{\mathbf{v}}_t^s) + \sigma_{t-2k} (\sigma_t \mathbf{z}_t - \alpha_t \hat{\mathbf{v}^s}_t)^s$
        
    \Else 
        \State $ \# \text{ 2 steps of DDIM with teacher}$
        \State $\mathbf{z}_{t-k} = \alpha_{t-k}(\alpha_t \mathbf{z}_t - \sigma_t \hat{\mathbf{v}}_t) + \sigma_{t-k} (\sigma_t \mathbf{z}_t -\alpha_t\hat{\mathbf{v}}_t)$
        \State $\mathbf{z}_{t-2k} = \alpha_{t-2k}(\alpha_{t-k} \mathbf{z}_{t-k} - \sigma_{t-k} \hat{\mathbf{v}}_{t-k}) + \sigma_{t-2k} (\sigma_{t-k} \mathbf{z}_{t-k} - \alpha_{t-k} \hat{\mathbf{v}}_{t-k})$

    \EndIf
    \State $t \leftarrow t-2k$
    \State  Get trajectory $\tau_i = (\mathbf{z}_t,\epsilon,t)$ based on induced distribution of $\mathbf{v}, \mathbf{v}^s$
    
\EndFor
\State Add $\tau_i$ to dataset, $\mathcal{D}_{DAgger} \leftarrow \mathcal{D}_{DAgger} \cup \tau_i$

\end{algorithmic}
\end{algorithm}

\begin{algorithm}
\small
\caption{DDIL: Progressive Distillation on the aggregated dataset and forward diffusion within DDIL framework, assumes PF-ODE and deterministic sampling.
}\label{alg:step_distill}\label{step distill}
\begin{algorithmic}
\Require Teacher diffusion models with text-conditioning and parameters $\theta$ 
\Require Data set $\mathcal{D}$
\Require Initialize DAgger dataset to collect trajectories, $ \mathcal{D}_{DAgger}  \leftarrow \emptyset $
\Require Number of teacher model denoising steps $N$
    
\For{$L$ iterations}
    \State $\eta \leftarrow \theta$ \Comment{Initialize student from teacher}
    \State $k = 1000/N$
    \While{not converged}
        \State $ \# \mathbf{z}_t \text{ from from aggregated dataset or forward process}$
        \If{$p \sim U[0,1] < p$} 
            \State $(\mathbf{z}_t,\epsilon,t) \sim D_{DAgger}$ \Comment{sampled from mixed unrolling \ref{alg:dagger}} 
        \Else 
            \State $\mathbf{z}_t = \alpha_t \mathbf{x} + \sigma_t \epsilon$, where $x \sim D$, $t \sim U[0,1]*k$, and $\epsilon \sim \mathcal{N}(\mathbf{0},\mathbf{I})$ \Comment{Forward process}
              
        \EndIf
     
        \State  $\hat{\mathbf{x}}_t^{s} = \alpha_t \mathbf{z}_t - \sigma_t \mathbf{v}_t^s$
        
        \State $ \# \text{ 2 steps of DDIM with teacher}$
        \State $\mathbf{z}_{t-k}^* = \alpha_{t-k}(\alpha_t \mathbf{z}_t - \sigma_t \hat{\mathbf{v}}_t^*) + \sigma_{t-k} (\sigma_t \mathbf{z}_t - \alpha_t \hat{\mathbf{v}}_t^*)$
        \State $\mathbf{z}_{t-2k}^* = \alpha_{t-2k}(\alpha_{t-k} \mathbf{z}_{t-k}^* - \sigma_{t-k} \hat{\mathbf{v}}_{t-k}^*) + \sigma_{t-2k} (\sigma_{t-k} \mathbf{z}_{t-k}^* - \alpha_{t-k} \hat{\mathbf{v}}_{t-k}^*)$
        
        \State $\hat{\mathbf{x}}_t^{Target} \equiv \hat{\mathbf{x}}_t =  \frac{\mathbf{z}_{t-2k}^T- \frac{\sigma_{t-2k}}{\sigma_t}\mathbf{z}_t}{\alpha_{t-2k}-\frac{\sigma_{t-2k}}{\sigma_t}\alpha_t}$ \Comment{Target Estimate}
        \State $L_{\eta} = \max\left(1,\frac{\alpha_t^2}{\sigma_t^2}\right)|| \hat{\mathbf{x}}_t^s(\eta) - \hat{\mathbf{x}}_t^{Target}||_2^2$
        \State $\eta = \eta- \gamma \nabla_{\eta} L_{{\eta}}$ \Comment{Optimization}

        \State Update $\mathcal{D}_{DAgger}$ using \Cref{alg:dagger} 
    \EndWhile
    \State $\theta \leftarrow \eta$ \Comment{Update teacher with current student}
    \State $N \leftarrow N/2$ \Comment{Halve the number of teacher denoising iterations}
    
\EndFor

\end{algorithmic}
\end{algorithm}

\section{Co-variate shift Analysis}

To validate our hypothesis of covariate shift from accumulation of error, we conduct a mixed-rollout evaluation using a 32-step CFG teacher and a 4-step DDIL-distilled student on the MS-COCO 2017 (5K) dataset. Both models achieve similar FID scores (~22.5) with the teacher model having a CLIP score of 0.321. We assume a 32-step teacher model, where 1 student step is equivalent to 8 teacher steps. This allows for alignment between teacher and student estimates at specific timesteps in the diffusion process ${999,749,500,250}$. This setting enables stochastic mixing between the teacher and student models during inference by choosing a state-less DDIM solver. 
We investigate three settings to assess if the teacher model can improve student generation from intermediate time steps. 
We vary the prior probability $(p_T)$ of selecting the teacher model for each (t-250) of transition. Results in Tab.~\ref{tab:selection_rate} show that decreasing $p_T$ (less frequent teacher usage) leads to a decline in CLIP score, suggesting the teacher model improves student predictions across various time steps.
To understand if there are any critical timesteps for teacher intervention, we analyze the impact of switching between teacher and student at specific time steps (Tab.~\ref{tab:teach2stud} and \ref{tab:stud2teach}). The results indicate that teacher intervention, either early or late in the diffusion process, can improve generation quality compared to student-only inference. This further supports the presence of covariate shift and its impact on student model.

\begin{table}[t!]
\vspace{-0.5 cm}
\scriptsize
\centering
\begin{minipage}[t]{.45\textwidth}
\caption{\small Evaluation with different teacher selection rates.} \label{tab:selection_rate}
\begin{tabularx}{\textwidth}{@{}Xcc@{}} 
\toprule

   \textbf{$p_{T}$} & \textbf{FID [$\downarrow$]} & \textbf{CLIP [$\uparrow$]}  \\ 
\midrule
                  0.8   & 23.14  &  0.319 \\
                  0.6   & 22.33  &  0.317 \\
                  0.4   &  21.95 & 0.313  \\
                  0.2   &  21.92 &  0.307 \\
                   0.0 & 22.42 & 0.302 \\
\bottomrule
\end{tabularx}
\vspace{0.7em}
\end{minipage}

\hfill
\begin{minipage}[t]{.45\textwidth}
\caption{\small Switching from $Teacher \to student$ within single generation.} \label{tab:teach2stud}
\begin{tabularx}{\textwidth}{@{}Xcc@{}} 
\toprule

   \textbf{${T}$} & \textbf{FID [$\downarrow$]} & \textbf{CLIP [$\uparrow$]}  \\ 
\midrule
                    749    &  23.60 &  0.309 \\
                    500     & 23.80  &  0.317 \\
                    250    &  24.21 &  0.320  \\
\bottomrule
\end{tabularx}
\vspace{0.7em}
\end{minipage}

\hfill
\begin{minipage}[t]{.45\textwidth}
\caption{\small Switching from $Student \to teacher$ within single generation.} \label{tab:stud2teach}
\begin{tabularx}{\textwidth}{@{}Xcc@{}} 
\toprule

   \textbf{${T}$} & \textbf{FID [$\downarrow$]} & \textbf{CLIP [$\uparrow$]}  \\ 
\midrule
749    &  21.64 &  0.321 \\
500     &  21.16 & 0.314  \\
250    &  22.05 &  0.306  \\
\bottomrule
\end{tabularx}
\end{minipage}
\label{tab:dagger_eval}
\vspace{-0.7em}
\end{table}


\subsection{Covariate Shift Visualization}

\begin{figure*}[h]
    \centering
    \includegraphics[width=1.0\textwidth]{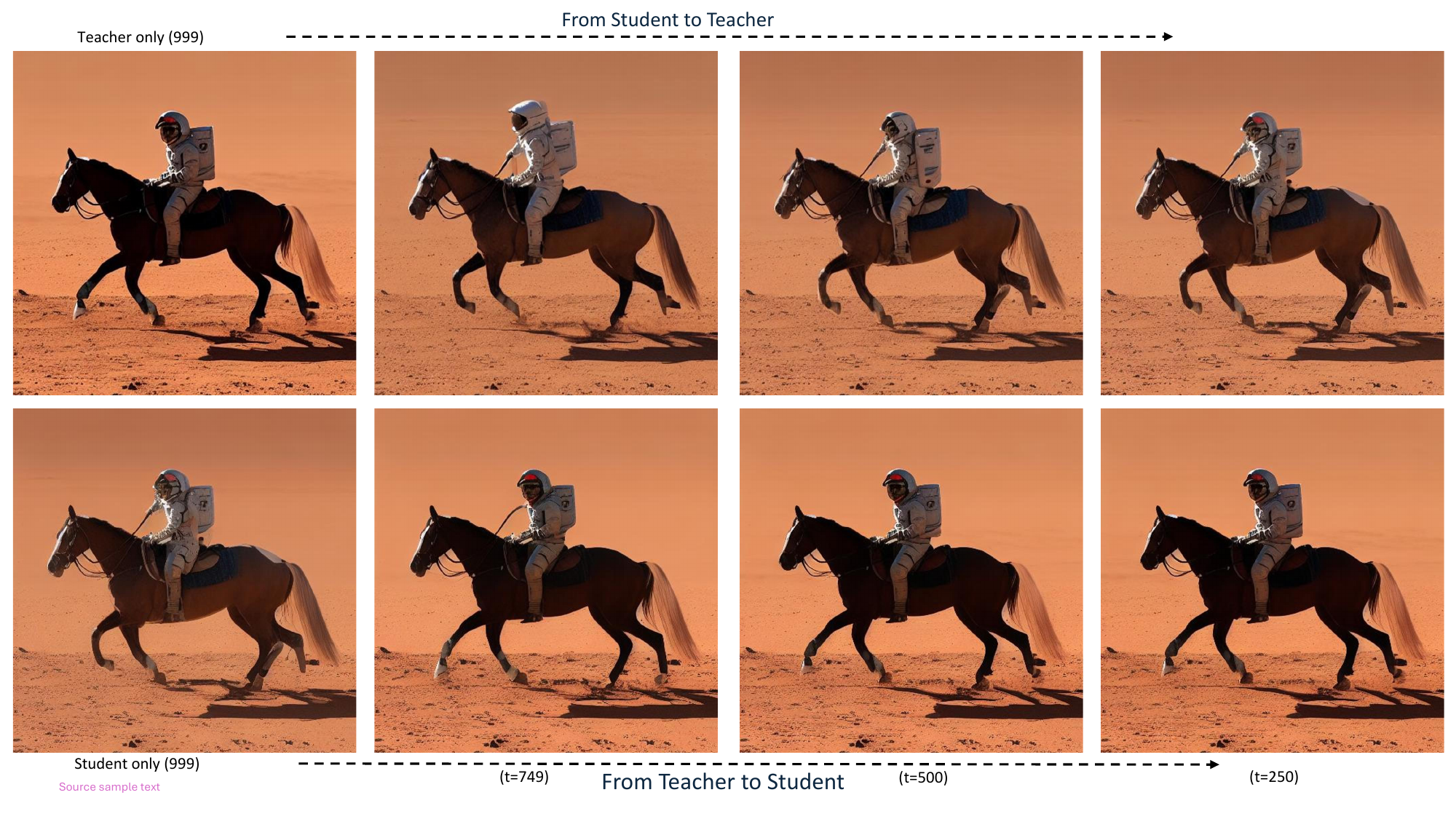}
    \caption{Sensitivity of timestep in reverse process}
    \label{fig:covariate_shift}
\end{figure*}

From Fig. \ref{fig:covariate_shift}, we can observe that more we use teacher model structural coherence is improved of the horse. More specifically if we switch from teacher to student from 750 i.e., run teacher model to provide initialization then run student model the structure is largely preserved. So this experiment shows importance of early denoising steps compared to later iterations within denoising process.

Alternatively if we switch from student to teacher i.e., see if the teacher has ability to correct and generate good image, we can observe from Fig \ref{fig:covariate_shift} if we switch to teacher after t=500 i.e, half of denoising steps teacher cannot fix already in-coherent structure of horse's legs (back). This experiment also \textbf{illustrates potential for non-uniform denoising for improved quality vs latency tradeoff} in diffusion models.

\begin{table*}\label{tab:prompt_inv}
\caption{Evaluating baseline based optimized/PH2P prompts on distilled models, showing the effectiveness of map-preserving multi-step distilled methods over other fewer step distillation methods}
\scriptsize
\begin{tabularx}{\textwidth}{@{}Xcccc@{}}
\toprule
Model             & Steps & LPIPS\_Div &  &  \\
\midrule
ADD               & 1     & 0.51       &  &  \\
LCM               & 4     & 0.45       &  &  \\
LCM-LoRA          & 4     & 0.53       &  &  \\
Instaflow(0.9B)   & 1     & 0.60       &  &  \\
SD                & 32    & 0.63       &  &  \\
Step Distillation & 8     & 0.61       &  &  \\
DDIL              & 8     & 0.60       &  &  \\
Step Distillation & 4     & 0.60       &  &  \\
DDIL              & 4     & 0.60       &  & \\
\midrule
\end{tabularx}
\label{tab:prompt_inv}
\end{table*}

\section{Prompt Inversion}

In \Cref{tab:prompt_inv}, we investigate if the underlying map from `noise' space to `data' space is preserved during the distillation phase of diffusion. 
Updating the map could have an implication on adopting various tools obtained on pre-trained diffusion model with applications in personalization etc. 
To capture this we consider a setting where we obtain inverted prompts of a reference image using \cite{mahajan2023prompting} on the COCO dataset and then pass the inverted prompts to distilled models to capture the similarity of the generated image to the reference image and also diversity of generations. 
As PH2P returns the optimal token for a given image, if a relative change in the map is minimal we expect the generated output to be more aligned with the reference image. If distilled model has good behavior we expect distilled models to preserve diversity of generation on invertd prompts too, our overall findings are consistent with text-guided generation for inverted prompts too.




\section{Additional Qualitative Examples}

In this section, we compare various trajectory based distillation techniques on SDv1.5 and show efficacy of DDIL(+PD) in terms of \textbf{quality, diversity} compared to other publicly available distilled checkpoints from Instaflow, ADD (SDv2.1) and LCM compared to pre-trained diffusion model SDv1.5.

\begin{figure*}[h]
  \centering
  \begin{minipage}{1.0\textwidth}
    \includegraphics[width=\linewidth]{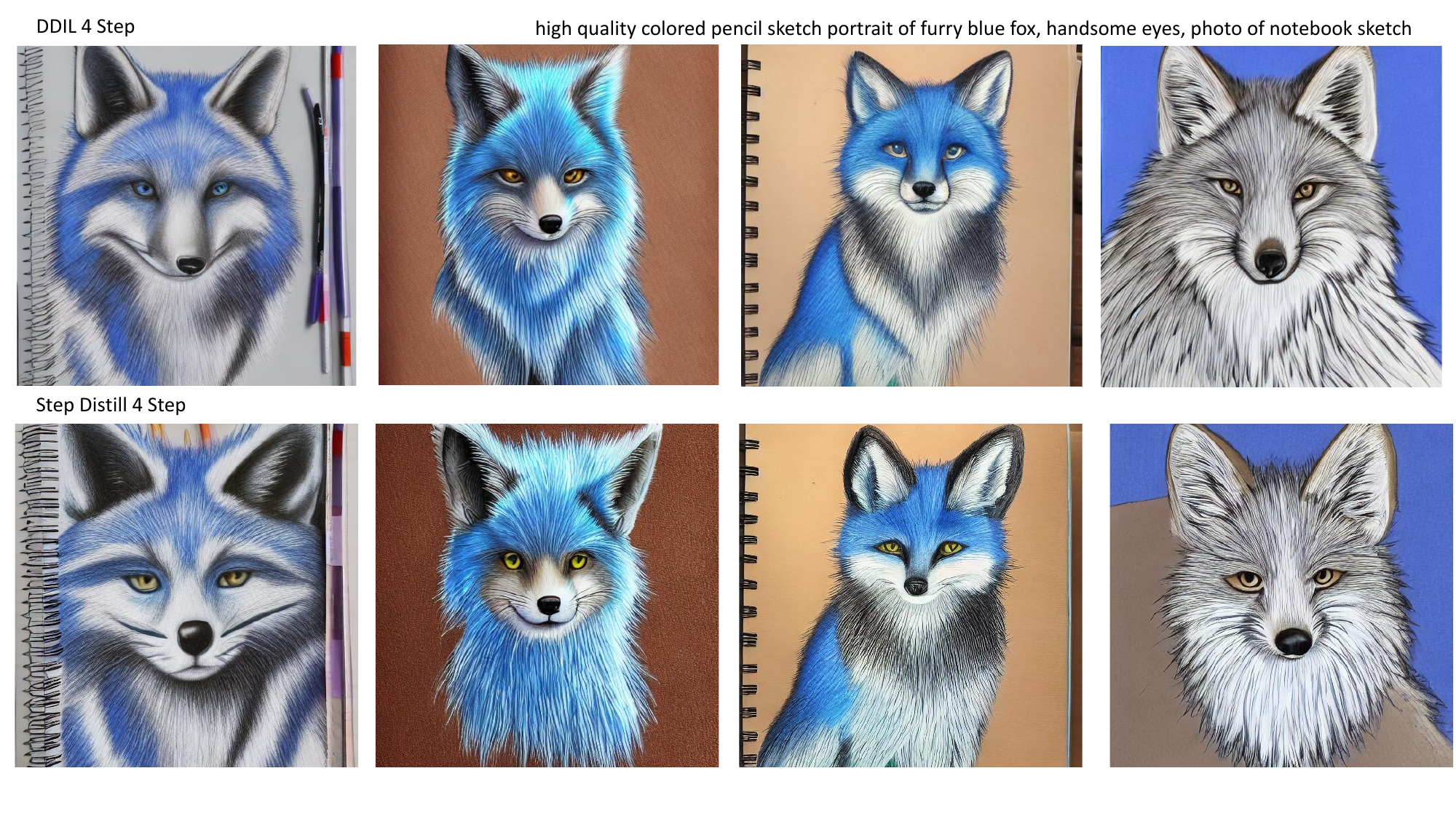}
    \label{fig:image1}
  \end{minipage}
  \begin{minipage}{1.0\textwidth}
    \includegraphics[width=\linewidth]{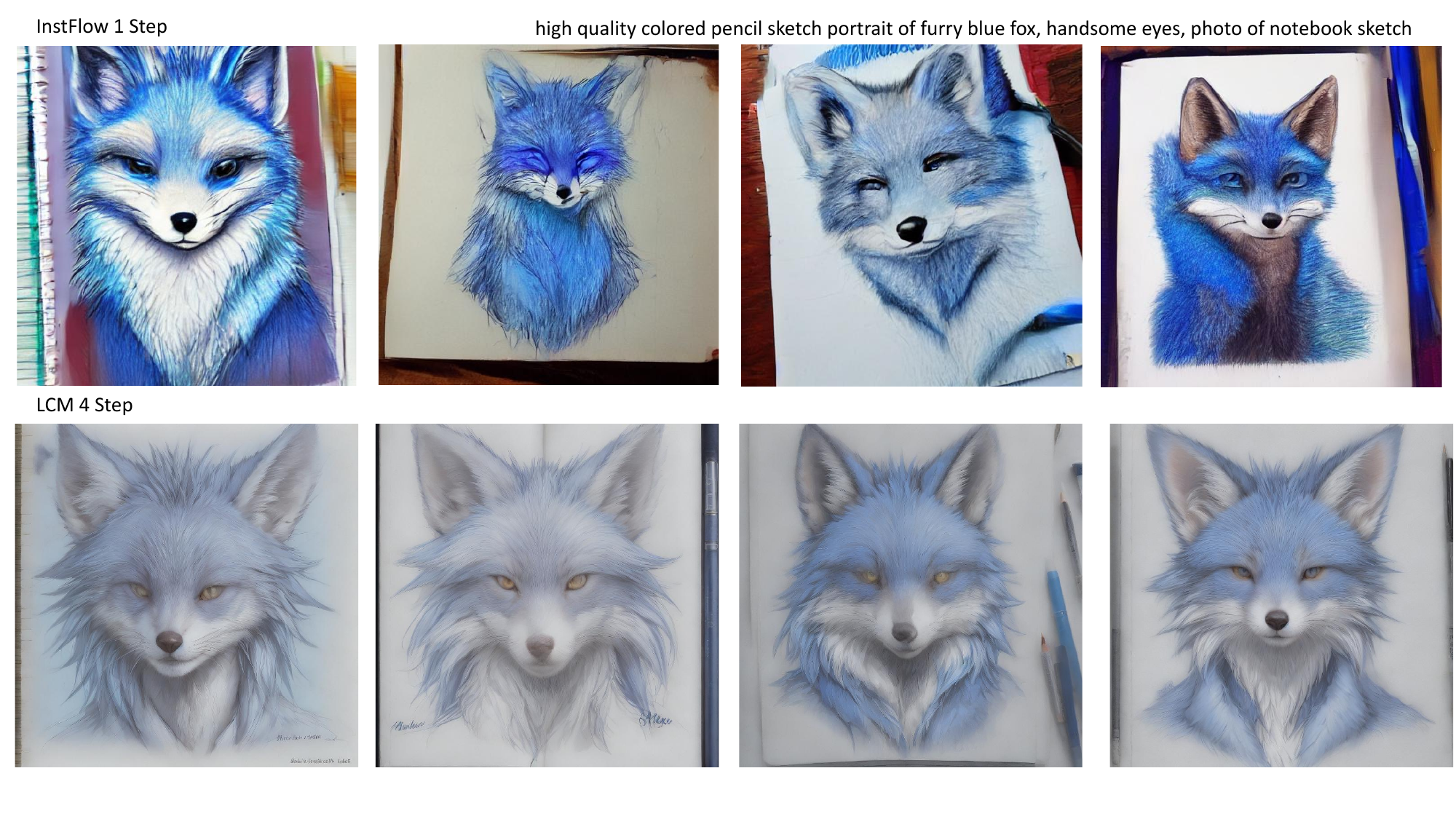}
    \label{fig:image2}
  \end{minipage}
  \hfill
\end{figure*}

\begin{figure*}[h]
  \centering
  \begin{minipage}{1.0\textwidth}
    \includegraphics[width=\linewidth]{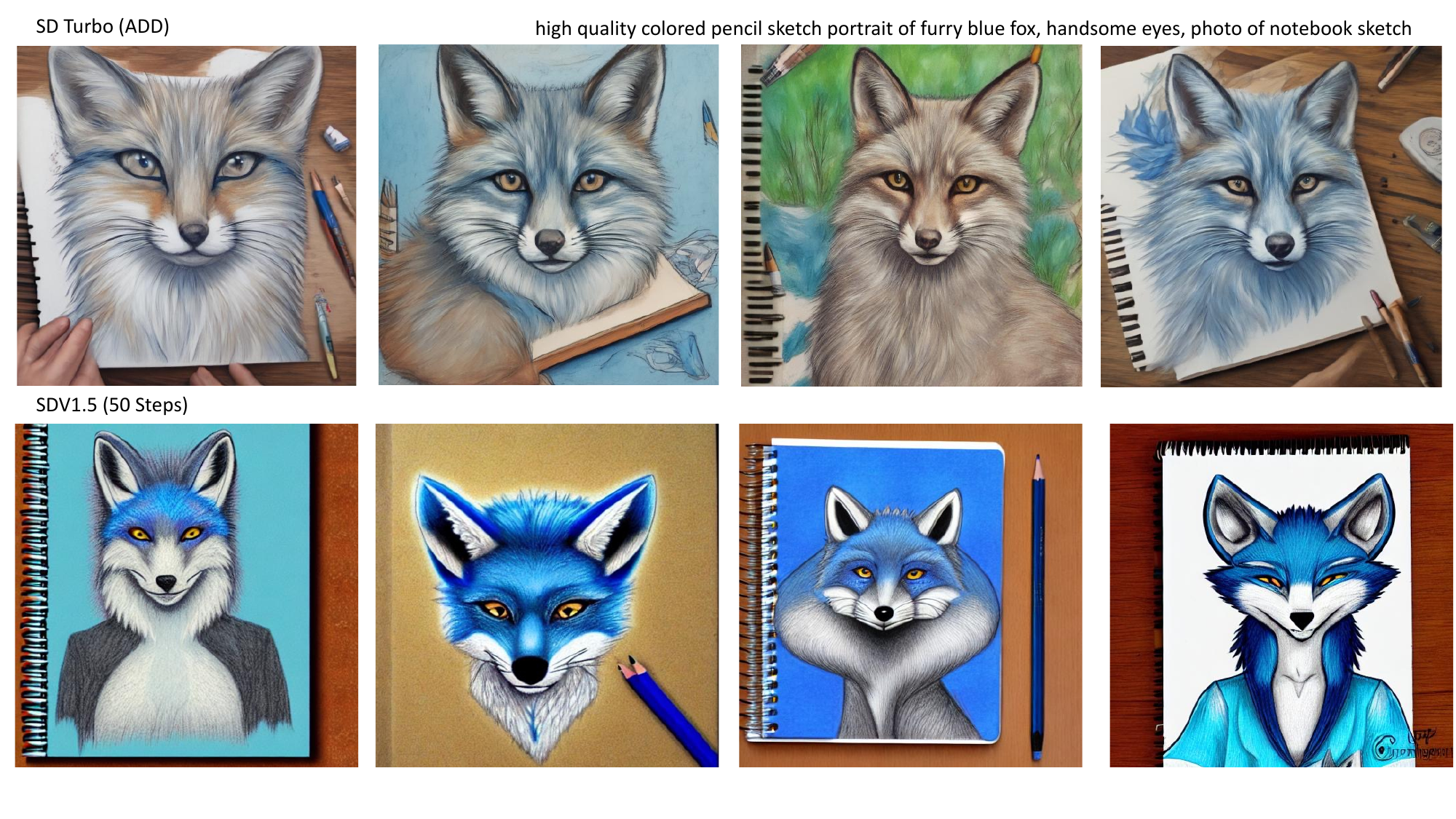}
    \label{fig:image1}
  \end{minipage}
  \hfill
\end{figure*}






